\pdfoutput=1

\documentclass[11pt]{article}

\usepackage[final]{neurips_2023}
\usepackage{natbib}
\setcitestyle{numbers}

\usepackage{amsmath}
\usepackage{amssymb}
\usepackage{mathtools}
\usepackage{amsthm}

\usepackage{microtype}
\usepackage{graphicx}

\usepackage{subcaption}
\usepackage{booktabs} 
\usepackage{algorithm}
\usepackage{algorithmic}
\usepackage[utf8]{inputenc} 
\usepackage[T1]{fontenc}    
\usepackage{hyperref}       
\usepackage{url}            
\usepackage{booktabs}       
\usepackage{amsfonts}       
\usepackage{nicefrac}       
\usepackage{microtype}      
\usepackage{xcolor}
\usepackage{array}
\usepackage{tabulary}
\newcolumntype{K}[1]{>{\centering\arraybackslash}p{#1}}
\usepackage{inconsolata}
\usepackage{hyperref}

\usepackage{times}
\usepackage{latexsym}

\usepackage{microtype}

\author{Jiachen Zhao\\
University of Massachusetts Amherst\\
{\tt jiachenzhao@umass.edu}}

\begin{document}
\title{Student as an Inherent Denoiser of Noisy Teacher}
\maketitle

\begin{abstract}
Knowledge distillation (KD) has been widely employed to transfer knowledge from a large language model (LLM) to a specialized model in low-data regimes through pseudo label learning. However, pseudo labels generated by teacher models are usually noisy and may influence KD performance. This study delves into KD with noisy teachers and uncovers that the student model can already generate more accurate predictions than the teacher labels used to train it during KD, indicating its inherent ability to \textit{denoise} noisy teacher labels. Motivated by this finding, we propose Peer-Advised KD to improve vanilla KD from noisy teachers. Experiments show that Peer-Advised KD can outperform LLM by approximately 5\% with 50 human-labeled data, and even competitive to standard supervised finetuning with 750 human-labeled data.
\end{abstract}

\section{Introduction}


Recently, with the emergence of large language models (LLMs), knowledge distillation (KD) with label matching has been widely applied to low-data regimes~\citep{shridhar2022distilling,ho2022large,li2023feasibility,gilardi2023chatgpt,wang-etal-2021-want-reduce,yoo-etal-2021-gpt3mix-leveraging,ding2022gpt}. In scenarios where there is a scarcity of labeled in-domain data but an abundance of unlabeled data, LLMs can serve as an annotator to generate pseudo labels through few-shot prompting, i.e., in-context learning~\cite{brown2020language}. These pseudo labels are then used to train a smaller student model through KD. The student model can offer improved efficiency and faster inference for deployment.  However, one crucial problem with such use of KD is that the generated pseudo labels of teacher models are usually noisy. The performance of student models can be greatly dependent on the quality of teacher labels~\cite{NEURIPS2022_2e343555}. Despite many past works on leveraging LLMs as annotators, few works have investigated how to optimize learning noisy data.  Some previous works for noisy label learning utilize loss reweighting~\cite{hinton2015distilling,NEURIPS2022_2e343555,lee2013pseudo} that may not apply to few-shot cases where pseudo labels are overwhelmingly more than gold labels, leading to great imbalance and unstable training.  
 
Therefore, this work aims at providing new insights into learning noisy teacher labels for more reliable student training in low-data regimes. This work mainly focuses on learning linguistic structures as our test bed, which usually requires extensive labeled data to acquire~\citep{drozdov-etal-2019-unsupervised-latent} and is the fundamental ability to downstream NLP tasks~\citep{manning2020emergent}.  More specifically, we first analyze the learning process of KD (Sec.~\ref{sec:theo}) with according empirical results (Sec.~\ref{sec:exp1}).  Our analysis reveals that \textbf{during KD, the student model exhibits a \textit{denoising} ability, generating superior predictions compared to the noisy teacher labels used to train it}.  Our findings imply that vanilla KD from noisy teacher gives sub-optimal student's performance.  Motivated by our findings, we present Peer-Advised KD (PA-KD) in Sec.~\ref{sec:main} which harnesses the student's inherent denoising ability to generate improved pseudo labels for KD. 



\section{Revisiting KD}


\paragraph{Definition 1} (KD for Semi-Supervised Learning) \textit{We assume there is an accessible set of labeled data $\mathcal{D}_{(x,y)}^{\text{labeled}}$ whose distribution is $p^{\ast}$. Then, given a set of unlabeled data $\mathcal{D}_{x}^{\text{unlabeled}}$ sampled from the same distribution $p^{\ast}$, $\mathrm{T}$ is employed to generate the distillation set $\mathcal{D}_{(x,\tilde{y})}^{\text{pseudo}}$.  The student model $\mathrm{S}$ is then trained by minimizing the loss $\mathcal{L} = \beta \ell(\hat{y},y) + \ell(\hat{y},\tilde{y})$.}

We first detail the setting of KD in our analysis. The student model is denoted by $\mathrm{S}$ whose label is $\hat{y}$, and the teacher model is denoted by $\mathrm{T}$ whose label is $\tilde{y}$. The teacher model $\mathrm{T}$ may leverage $\mathcal{D}_{(x,y)}^{\text{labeled}}$ through In-Context Learning~\citep{brown2020language} or supervised finetuning to generate pseudo labels.  Additionally, we set $\beta$=0 to focus on the effects of distillation set following~\cite{stanton2021does}.  

\subsection{Decoding the Learning Process in KD}
\label{sec:theo}
This section analyzes the learning process during KD theoretically.  We hypothesize that at the beginning of KD,  the student $\mathrm{S}$ will first fully converge to partial pseudo labels of teacher $\mathrm{T}$ (the set of which is denoted as {$Y_{learnt}$}, while the corresponding $x$ is in {$X_{learnt}$}). Because the learning process is usually imbalanced where the weights are updated more frequently for information that occurs more often in the training dataset~\cite{sun2009classification,liu2019large,cui2019class,wang2017learning}. 
 Formally, after $t_{0}$ training steps, we consider $\mathrm{S}$ has learnt a set of knowledge denoted by $\xi_{base}$ and has nearly zero loss on $Y_{learnt}$. At the next training step, we can denote $\hat{y}= \mathrm{S}(x|\xi_{base})$ which means the student's prediction is conditioned on its learnt knowledge $\xi_{base}$. Accordingly, we represent teacher's encoded knowledge with $\xi_{\mathrm{T}}$.  Then, for the next incoming training datapoint $(x,\tilde{y})$, we consider two cases depending on underlying patterns of $x$ and analyze respective learning effects. We focus on the comparison between student's predictions and teacher labels to learn. Formally, we define $\delta:=s({y},\mathrm{S}(x|\xi_{base}))-s({y},\tilde{y})$ where $s$ is a score function (F1 score for parsing) for similarity measurement.

\paragraph{Case 1: the underlying patterns of $x$ are the same as or similar to patterns of instances in $X_{learnt}$.}\label{sec:theo-1}  (1) If $\tilde{y}\in Y_{learnt}$, i.e., teacher labels that $\mathrm{S}$ has converged to, then we can get $\delta = 0$. (2) If $\tilde{y} \notin Y_{learnt}$, we speculate that $\delta$ can be unclear. $\delta$ may be positive (i.e., $s({y},\mathrm{S}(x|\xi_{base}))>s({y},\tilde{y})$), indicating the student generates better labels than the teacher.  This is because $\xi_{base}$ might be much richer than $\xi_{\mathrm{T}}$ for the specific pattern of $x$, especially when $|\mathcal{D}_{x}^{\text{unlabeled}}| >>|\mathcal{D}_{(x,y)}^{labeled}|$.  Chances are that $(X_{learnt},Y_{learnt})$ after some $t_{0}$ may already contain extensive credible information about diverse cases similar to $x$, contributing to strong $\xi_{base}$ which is conditioned on by $\mathrm{S}$ for its prediction.  In this scenario during KD, the student's performance may be degraded because it is forced to learn inferior teacher labels to its own predictions.  We empirically demonstrate this scenario in Sec.~\ref{sec:exp1}, which we refer to as student's inherent denoising ability.




\paragraph{Case 2: the underlying patterns of $x$ are clearly distinct from patterns of instances in $X_{learnt}$.}\label{sec:theo-2} (1) $\delta$ may tend to be negative indicating that the teacher labels are more correct ((i.e., $s({y},\mathrm{S}(x|\xi_{base}))<s({y},\tilde{y})$)). Because knowledge of $\xi_{base}$ involving patterns of $x$ may have not been learnt effectively by $\mathrm{S}$ after $t_{0}$. In this case, $\mathrm{S}$ may benefit from KD by learning new knowledge.  (2) If patterns of $x$ are uncommon, $\delta$ can be unclear because both student and teacher may be making educated guesses conditioned on learnt knowledge.  Chances are that both predictions are not credible.
 

All in all, we hypothesize that KD is an iterative process of alternating between case 1 and case 2 discussed above. This analysis of KD can be viewed as ``expansion''~\cite{wei2021theoretical} from the learnt information to the under-learnt. 

%

\subsection{Student Can Inherently Denoise Teacher Labels}
\label{sec:exp1}
\begin{figure}[t]
    \centering
    \begin{subfigure}{0.32\textwidth}
        \includegraphics[width=\linewidth]{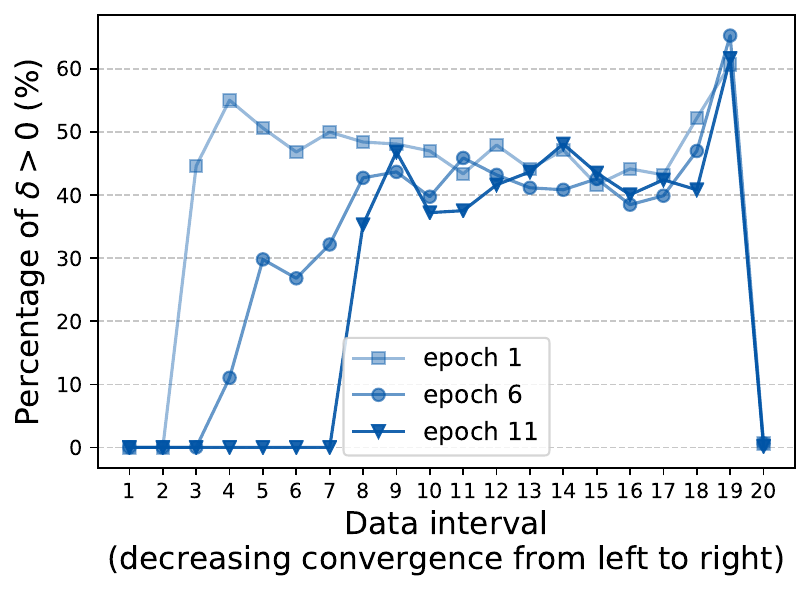}
        \caption{KD w/ LLM on CRAFT}
    \label{fig:denoise_ptb}
    \end{subfigure}
    \hfill
    \begin{subfigure}{0.32\textwidth}
        \includegraphics[width=\linewidth]{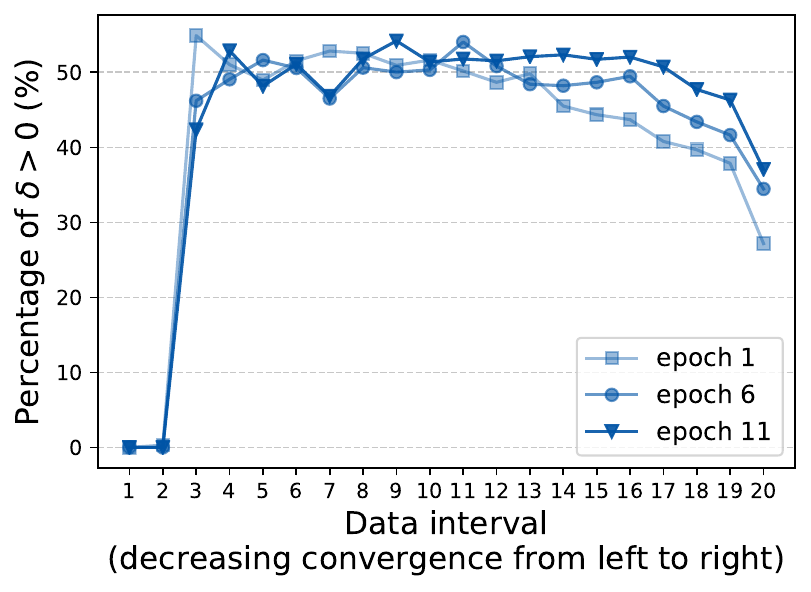}
        \caption{KD w/ LLM on PTB}
    \label{fig:denoise_craft}
    \end{subfigure}
    \hfill
    \begin{subfigure}{0.32\textwidth}
        \includegraphics[width=\linewidth]{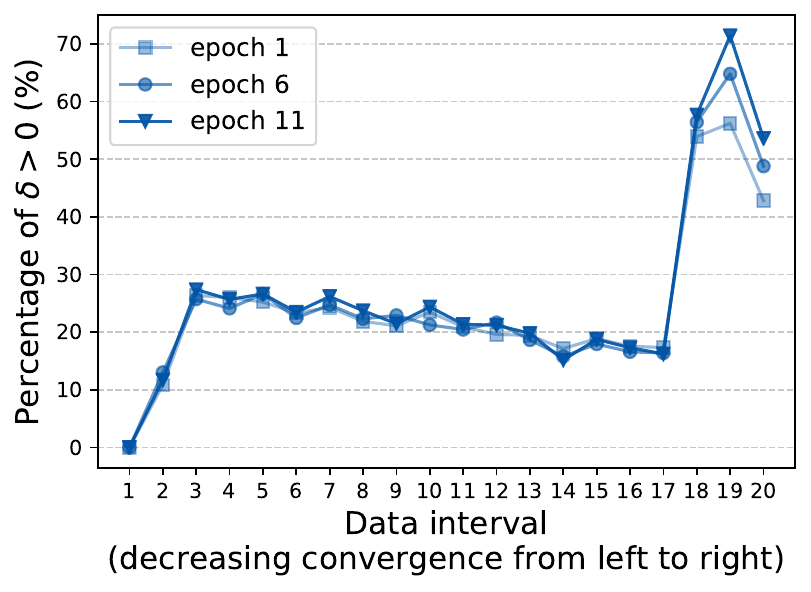}
        \caption{KD w/ supervised teacher}
    \label{fig:denoise_self}
    \end{subfigure}
    \caption{During KD, the student can actually generate better labels than the teacher labels used to train it. The y-axis is percentage that the student's predicted labels are better than teacher's pseudo labels for training (see definition for $\delta$ in Sec.~\ref{sec:theo}). The x-axis is the same as Fig.~\ref{fig:converge}.}
\label{fig:denoise}
\end{figure}

In this section, experiments are conducted to provide empirical understandings of KD.  We employ two datasets Penn Tree Bank and CRAFT. Teacher models are respectively Codex~\cite{chen2021evaluating} and T5-base~\cite{raffel2020exploring}.  F1 score is used to compare the similarity of labels (i.e., parsed trees).  Detailed implementations are in Appendix~\ref{app:exp_set_1}.  During knowledge distillation (KD), the student's predictions on training data are compared with the teacher labels used for training it.  Notably, we observe that the student demonstrates a denoising ability, generating better labels than some teachers labels used to train it in the middle of KD as shown in Fig.~\ref{fig:denoise}.  This implies vanilla KD is sub-optimal since the student is forced to learn inferior teacher labels to its own predictions sometimes. 

\begin{figure}[t]
    \centering
    \begin{subfigure}{0.32\textwidth}
        \includegraphics[width=\linewidth]{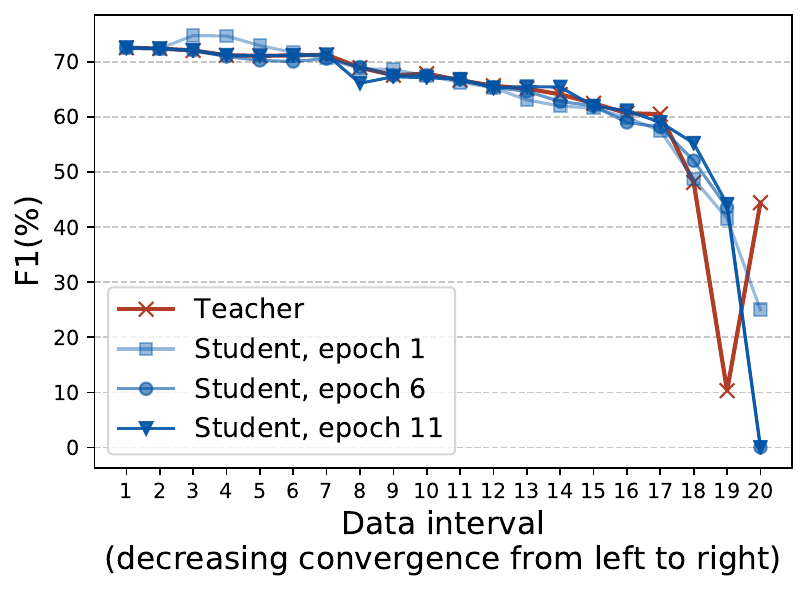}
        \caption{KD w/ LLM on CRAFT}
    \label{fig:converge_ptb}
    \end{subfigure}
    \hfill
    \begin{subfigure}{0.32\textwidth}
        \includegraphics[width=\linewidth]{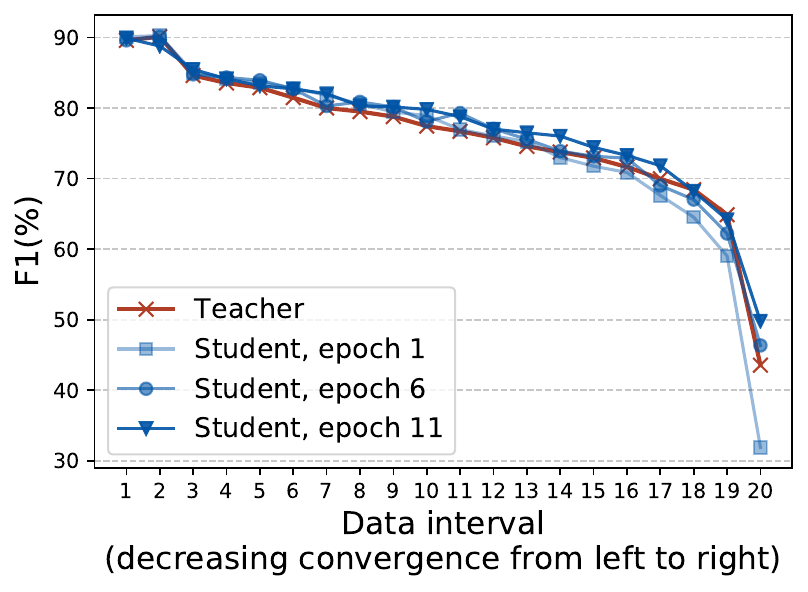}
        \caption{KD w/ LLM on PTB}
    \label{fig:converge_craft}
    \end{subfigure}
    \hfill
    \begin{subfigure}{0.32\textwidth}
        \includegraphics[width=\linewidth]{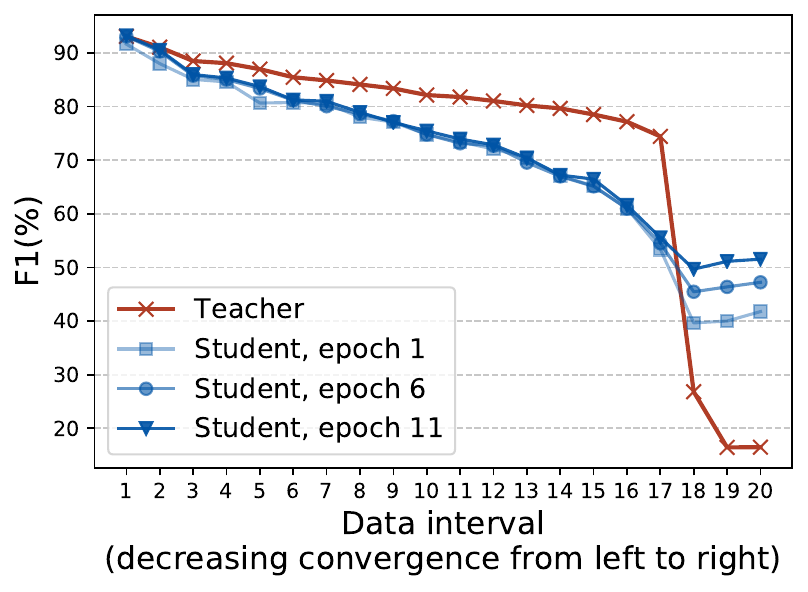}
        \caption{KD w/ supervised teacher}
    \label{fig:converge_ptb_self}
    \end{subfigure}
    \caption{The student model tends to converge more to less noisy pseudo labels of teacher during training. The y-axis is F1 w.r.t ground truth.  The x-axis is sets of data ranked in descending order based on student's convergence with teacher labels. Each interval range is 5\% of the size of dataset.  From Fig.~\ref{fig:converge_ptb} to Fig.~\ref{fig:converge_ptb_self}, the general noise level of teacher labels is decreasing.}
\label{fig:converge}
\end{figure}

 \paragraph{How does the denoising effect occur?}  We identify such denoising effect can be attributed to the discrepancies in learning teacher labels, as shown in Fig.~\ref{fig:converge}. 
Student converges faster to less noisy teacher labels than noisier ones. \citet{arpit2017closer} observed such discrepancies in adversarial label learning where labels are randomly flipped to introduce noises.  Our results suggest that the finding also holds for sequence-level KD where noises in labels are not random but generated by a teacher model, implying its inductive bias.  Further analysis is provided in Appendix~\ref{app:dis_converge}. 
 When examining Fig.~\ref{fig:converge_ptb} and Fig.~\ref{fig:denoise_ptb} (or Fig.~\ref{fig:converge_craft} and Fig.~\ref{fig:denoise_craft}, or Fig.~\ref{fig:converge_ptb_self} and Fig.~\ref{fig:denoise_self}), the student is observed to, after converging to more credible teacher labels (higher ground-truth F1), outperform the noisier teacher labels (lower ground-truth F1) by conditioning on its learnt knowledge.  This falls into the scenario (2) in the Case 1 of our theoretical framework of KD in Sec.~\ref{sec:theo-1}.  Due to student's inclination to converging to cleaner teacher labels, the student's learnt knowledge $\xi_{base}$ at some time in the middle of KD can be possibly reliable and even much more sufficient than teacher's $\xi_{\mathrm{T}}$ for some patterns, especially when $|\mathcal{D}_{x}^{\text{unlabeled}}| >>|\mathcal{D}_{(x,y)}^{labeled}|$. This is also evidenced by experiments in Appendix~\ref{app:size_distill} showing that denoising ability is more noticeable when the distillation set (i.e., noisy teacher's labels) is much larger than the real labels used to train the teacher model.  Further analysis between our theoretical framework of KD and empirical results are shown in Appendix~\ref{app:denoise}.


\section{Peer-Advised Knowledge Distillation from Noisy Teachers}\label{sec:main}
In this section, we propose Peer-Advised Knowledge Distillation (PA-KD) for learning from noisy teachers in low-data regimes motivated by our findings.  We have shown that during KD, the student can outperform the noisier teacher labels by conditioning on its learnt knowledge because the student tends to converge to more credible teacher labels faster.  To leverage and magnify this process in KD, we first extract teacher labels denoted by $\mathcal{T}_{\text{high}}$ based on convergence to train a peer student model $\mathrm{S}_{1}$, which is then employed to re-annotate the left teacher labels other than $\mathcal{T}_{\text{high}}$ to get peer labels $\mathcal{S}$.  More specifically, within dataset $\mathcal{T}:=\{(x,\tilde{y})\}$ and the set of all teacher labels $\tilde{Y}$, a partition function is defined as follow,
\begin{align}
    \mathcal{T}_{\text{high}}&=\{(x,\tilde{y}) | s(\tilde{y},\hat{y}) >= M(\tilde{Y},r)\},\label{exp:part1} 
\end{align}
where $M(\tilde{Y},r)$ stands for the threshold score of top $r$\% predictions among $\tilde{Y}$ and $s$ is F1 score function to measure the similarity between the student labels $\hat{y}$ and teacher labels $\tilde{y}$ as a proxy indicating convergence. 
 Finally, the ultimate student model $\mathrm{S}_{2}$ will be trained with $\mathcal{T}_{\text{high}}$ and $\mathcal{S}$ for improved performance. In addition, we propose another baseline, i.e., selective KD, which uses $\mathcal{T}_{\text{high}}$ alone for KD rather than using full data to reduce influence of noises in teacher labels.



\begin{table}[t]
\centering
\resizebox{\textwidth}{!}{
\renewcommand{\arraystretch}{1.3}
\begin{tabular}{llccK{2cm}K{2cm}K{2cm}K{2cm}}
\toprule
    & \multicolumn{2}{l}{}                    & \textbf{\#Params}    & \multicolumn{2}{c}{\textbf{PTB}}            & \multicolumn{2}{c}{\textbf{CRAFT}}          \\ \hline
\multicolumn{3}{l}{\textbf{\# Human-labeled Data}}                 &                      & 50                   & 250                  & 50                   & 250                  \\ \hline
\multicolumn{3}{l}{Teacher: Code-Davinci-002} &   175B                   &74.51                    &  76.49                    & 62.27                   &    64.37                  \\
\multicolumn{3}{l}{Student: T5-base}          &   220M                   &0                      & 58.02                     & 0                     &  54.11                    \\ \cline{3-8} 
    &                        & SLKD           &    220M                  &  77.21                    & 77.40                    & 65.58                     &    66.71                  \\
    &                        & SLKD+SD        &  220M                    &  77.34                    & 77.78                     & 67.21                     & 68.05                     \\
    &                        & SLKD+SD w/ HC        &  220M                    & 77.11                    & 78.22                     &  67.33                    & 68.15                     \\  \cline{3-8}
            &                        & Selective KD   &   220M                   &  78.13                    & 78.44                    & 68.24                     &    68.37\\
& &{PA-KD}    & 220M                     & \textbf{78.64}                     & \textbf{78.98}                     & \textbf{68.98}                     & \textbf{68.92}        \\                
\bottomrule
\end{tabular}
}
\caption{Performance (F1 score (\%)) of different methods on PTB and CRAFT in two few-shot settings.}
\label{tab:mainret}
\end{table}

\subsection{Experimental Results}
\label{sec:exp_main}

We utilize the Penn Treebank (PTB) corpus~\cite{marcus-etal-1993-building} and the Colorado Richly Annotated Full-Text (CRAFT) corpus~\cite{cohen2017colorado} in our experiments for grammar induction. More details on the dataset and implementations are shown in Appendix~\ref{app:data} and Appendix~\ref{app:exp_set_2} respectively.  We consider three baseslines for comparison with our proposed PA-KD. (1) SLKD: knowledge distillation by directly learning decoded sequences proposed by \citet{kim-rush-2016-sequence}, which is the default way to do KD throughout this paper. (2) SLKD+SD: after KD, the student is retrained through self-distillation (SD)~\citep{Furlanello2018BornAN,mobahi2020sd,liu-etal-2021-noisy}, where labels of the same training data generated by itself are employed. (3) SLKD+ SD w/HC: A variant of baseline 2 is to conduct SD only with labels that have high output confidence (HC) of the model, which is widely adopted to obtain better pseudo labels~\citep{vinyals2015grammar,wang2021selective,DBLP:conf/naacl/McCloskyCJ06,lang2022co}. 


Results are displayed in Table~\ref{tab:mainret}. With few-shot prompting, the LLM with limited labeled data can give decent performance. In comparison, the T5-base model trained with a small set of labeled data performs much worse than teacher. When there are 50 real data, T5 base cannot produce meaningful predictions.  However, when training T5 base with massive pseudo labels through KD, it can outperform the teacher. Furthermore, our Peer-Advised KD can significantly improve vanilla KD and surpass all baselines. For example, on PTB, PA-KD improves student's F1 by around 2\%.  Additionally, PA-KD with 50 labeled data can outperform finetuning the student model with 500 data (results shown in Fig.~\ref{fig:sft} in Appendix). PA-KD can thus greatly save labeling cost. On the other hand, Table~\ref{tab:mainret} demonstrates that selective KD using highly converged teacher labels achieves impressive performance, surpassing KD on the entire dataset. This suggests that vanilla KD on full data is sub-optimal due to the noise present in the teacher labels. By utilizing our finding that student tends to converge faster to cleaner teacher labels, noisier teacher labels may be filtered based on convergence and thus student can avoid learning misleading noise, resulting in improved KD performance but with much fewer data.   Our approach is also effective for supervised teacher and self-distillation (see Appendix~\ref{app:super} and Appendix~\ref{app:self}).

\section{Conclusion}
In this work, we investigate knowledge distillation from a noisy teacher (e.g., LLM) through pseudo label learning. Extensive experiments show that the student inclines to converge to cleaner teacher labels and can utilize its acquired knowledge to denoise noisy teacher labels during KD.  Motivated by our findings, we then propose a novel Peer-Advised KD approach which leverages highly-converged teacher labels to train a peer model to relabel under-converged teacher labels (that are possibly noisier).  Experiments show our method is label-efficient and can consistently outperform strong baselines on two public datasets in low-data regimes. 


\bibliographystyle{plainnat}
\bibliography{ref}

\newpage
\appendix

\begin{figure}
    \centering
    \includegraphics[scale=0.45]{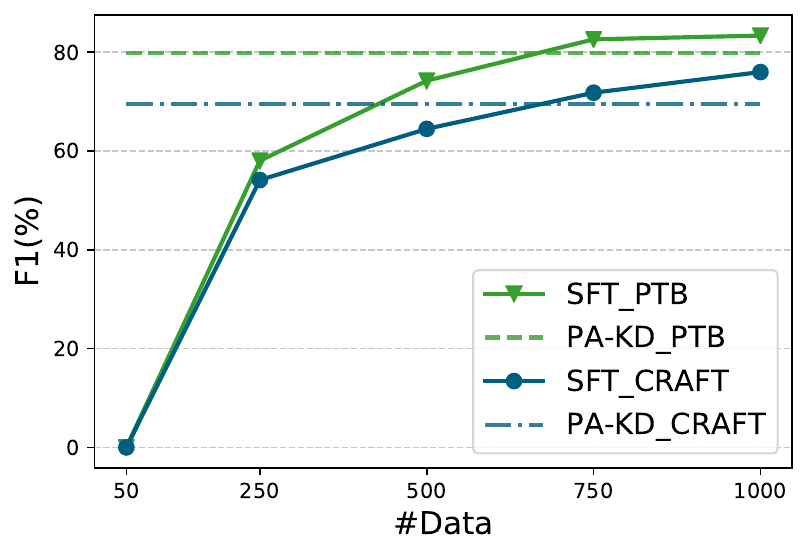}
    \caption{Comparison between Peer-Advised KD (PA-KD) with 50 labeled data and supervised-finetuning (SFT) T5-base with different amounts of labeled data. }
    \label{fig:sft}
\end{figure}

\begin{figure}
    \centering
    \includegraphics[scale=0.43]{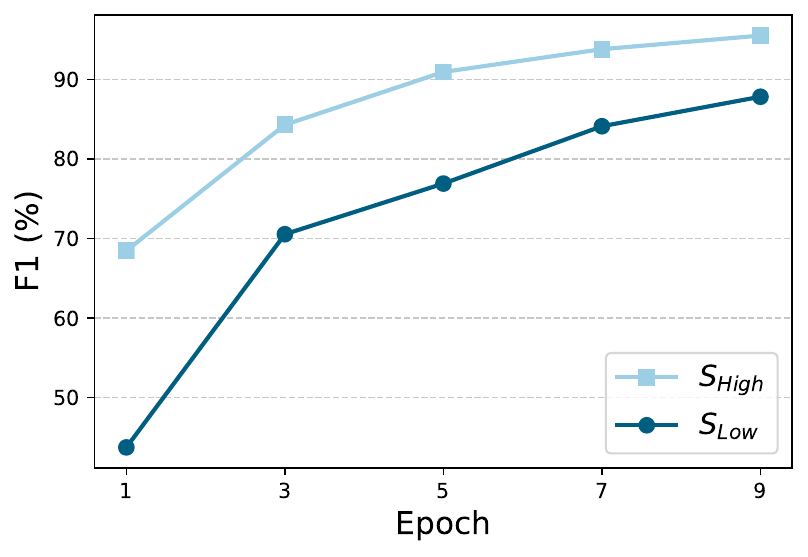}
    \caption{Convergence comparison (indicated by F1 w.r.t teacher labels) for students trained with good teacher labels ($\mathrm{S}_{High}$) and bad teacher labels ($\mathrm{S}_{Low}$). The division of teacher labels is binary and based on ground-truth F1 scores, e.g., the half with higher F1 w.r.t ground-truth is used to train $\mathrm{S}_{High}$. }
    \label{fig:disparity}
\end{figure}

\section{Discrepancy in Convergence}\label{app:dis_converge} 
We find during KD, the student demonstrates a greater inclination towards aligning with high-quality teacher labels. In all cases of Fig.~\ref{fig:converge}, as the quality of teacher labels (measured by ground-truth F1) decreases, we can observe a gradual reduction in the student's convergence with the teacher labels. Fig.~\ref{fig:disparity} clearly shows the student tends to converge faster to good teacher labels than bad ones. Such disparity in convergence has also been observed in other domains~\cite{dehghani2017fidelity,wang2019symmetric}. Thus, noise level of teacher labels can be estimated by convergence extent of student with teacher, which may be leveraged for label filtering. In terms of why less noisy teacher pseudo labels are more learnable, we speculate that the main reason is the inconsistency of errors in teacher labels.  The errors made by teacher can be arbitrary and instance-specific.  Such lack of consistent patterns across noisy labels prevents student from applying what it has learnt from one training sample to another, thus leading to slow convergence in learning.  

\begin{figure}[htp]
    \centering
        \includegraphics[scale=0.42]{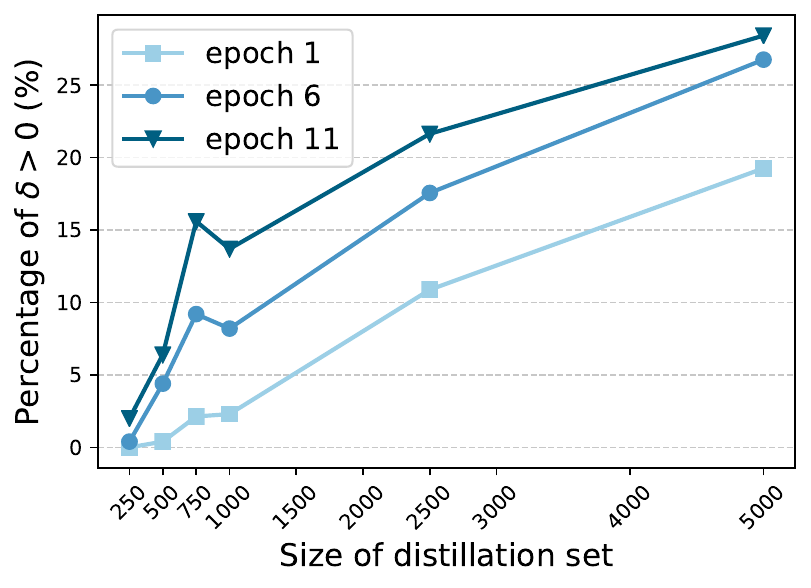}
        \caption{The figure shows the change of average percentage that student outperforms its teacher during KD with respect to the size of distillation set on PTB. In this case, the student is T5-small, while the teacher is T5-base trained with 500 real labels.}
    \label{fig:denoise_cmp}
\end{figure}

 \section{Denoising Effects and the Size of Distillation Sets}\label{app:size_distill}
 We identify that the emergence of denoising effect is related to the size of distillation set. As shown in Fig.~\ref{fig:denoise_cmp}, the average percentage that the student denoise teacher's labels will increase with more teacher labels used for training. When the number of teacher labels for student training is much larger than data (500 real labels) used to train teacher model, the student can have noticeable denoising effect despite the smaller size.  This may support our hypothesized situation (2) in Case 1 of Sec.~\ref{sec:theo-1} that sufficient teacher labels can contain more information than a small set of real labels.

\section{Further Analysis of Denoising Effects of Student}
\label{app:denoise}
Denoising effects may diminish as convergence decreases, particularly at the early epoch (see Fig.~\ref{fig:denoise_craft} and Fig.~\ref{fig:denoise_ptb}) and for the case that the teacher is trained through supervised tuning in Fig.~\ref{fig:denoise_self}).  We attribute this observation to the first scenario in Case 2 in Sec.~\ref{sec:theo-2}.  The reduced denoise effect of student can be caused by the increase of unfamiliar knowledge involved in samples that have low convergence between student and teacher labels. 
 Additionally, random spikes are observed in cases of extremely low convergence (see Fig.\ref{fig:denoise_craft} and Fig.\ref{fig:denoise_self}). In those cases, both the student and teacher actually exhibit low ground-truth F1 scores (see Fig.\ref{fig:converge_craft} and Fig.\ref{fig:converge_ptb_self}), indicating that they might be making educated guesses given some rare patterns or features in the input data. This may be an example of our second hypothesized scenario in Case 2 in Sec.~\ref{sec:theo-2}

\section{Experiments Details}
\subsection{Datasets}\label{app:data}
We process PTB dataset with the standard splits (2-21
for training containing 39832 sentences, 22 for validation, 23 for test). We randomly split the CRAFT corpus into a training set, a validation set and a test set following the ratio of 6:2:2. We consider two low-data regimes. We randomly select 50 and 250 samples in the training set as two cases of available annotated data, while the rest are treated as unlabeled data.

\subsection{Implementations for Experiments in Sec.~\ref{sec:exp1}}
\label{app:exp_set_1}
 We assume only 250 real labeled data available, while the rest of training data are considered as unlabeled. We employ Codex-Davinci-002 as teacher model and T5-base as student.  We use 50 exemplars for PTB and 35 exemplars for CRAFT in terms of prompting LLM.  On the other hand, for KD with a supervised teacher, the teacher model is T5-base and the student is T5-small. The teacher is fine-tuned with 500 in-domain data for 15 epochs on PTB. All supervised training entails a learning rate of $3\times10^{-4}$ and a batch size of 32.


\subsection{Implementations for Experiments in Sec.~\ref{sec:exp_main}}\label{app:exp_set_2}
In knowledge distillation, the student model is T5~\cite{raffel2020exploring} trained using hard labels, i.e., decoded sequences, also known as sequence-level KD~\cite{kim-rush-2016-sequence}.  For prompting, we utilize Codex (code-davinci-002) as the teacher model which is shown effective for parsing~\cite{DBLP:conf/emnlp/ShinLTCRPPKED21,DBLP:conf/naacl/ShinD22}. To avoid the number of demonstrations from surpassing token limits, we use 50 exemplars for PTB and 35 exemplars for CRAFT. Supervised training entails a learning rate of $3\times10^{-4}$, a batch size of 32, and 20 epochs. For convergence-based partition in Peer-Advised KD, an initial student $\mathrm{S}_{0}$ is trained with 2 epochs whose predictions on training data are compared with teacher labels for deciding convergence. $r$ is set as 50\%, i.e., the half of teacher labels with greater convergence with student's predictions are considered as $\mathcal{T}_{\text{high}}$.  For SD w/ HC, weights with highest validation score is used to generated labels, while for sole SD, student is trained for 4 epochs to avoid generating the same labels to teacher labels due to overfitting.  We set the half with above-mean output probabilities as HC labels.  During evaluation, sentence-level unlabeled parsing F1 scores are reported. These scores are computed individually for each sentence and then averaged across the dataset, aligning with previous works on parsing~\cite{DBLP:conf/emnlp/ShiLG20}.


\section{Results on Peer-Adivsed KD with Supervised Teacher}
\label{app:super}
Our proposed peer-advised KD approach also works for the supervised teacher.  We use T5-base as teacher model and T5-small as the student and assum access to 500 human-labeled data. Results are shown in Table~\ref{tab:superkd}. 
 In this case, a much smaller student (T5-small) trained with real data can only reach 21.92\% F1 on PTB, which is not comparable to the teacher, T5-base. KD can still significantly improve T5-small's performance, but the student still performs inferior to the teacher due to incapacity. However, our peer-advised KD surpass the vanilla KD by 1.94\%, enabling the student to be competitive to the teacher.
\begin{table}[htp]
\centering
\begin{tabular}{ccclcl}
\toprule
\multicolumn{2}{c}{Methods}  & \multicolumn{2}{c}{\#Params} & \multicolumn{2}{c}{F1} \\ 
\hline

&Teacher: T5-base     & \multicolumn{2}{c}{220M}   & \multicolumn{2}{c}{74.21}                                       \\
&Student: T5-small    & \multicolumn{2}{c}{60M}     & \multicolumn{2}{c}{21.92}                                                 \\ \cline{2-6} 
&SLKD       & \multicolumn{2}{c}{60M}  & \multicolumn{2}{c}{73.05}                                                \\
&SLKD+SD       & \multicolumn{2}{c}{60M} & \multicolumn{2}{c}{74.12}                                                \\
&SLKD+SD w/HC      & \multicolumn{2}{c}{60M}  & \multicolumn{2}{c}{71.17}                                                  \\\cline{2-6} 
&Selective KD      & \multicolumn{2}{c}{60M}   & \multicolumn{2}{c}{73.72}                                                   \\
&PA-KD    & \multicolumn{2}{c}{60M} &\multicolumn{2}{c}{74.60}                                                  \\
\bottomrule
\end{tabular}
\caption{Experimental results on knowledge distillation from a supervised-trained teacher.}
\label{tab:superkd}
\end{table}

\section{Results on Peer-Adivsed KD for Self-distillation}\label{app:self}
\begin{table}[htp]
\centering
\begin{tabular}{cccccc}
\toprule
\multicolumn{2}{c}{Methods} &   & \multicolumn{2}{c}{F1} \\ \hline
&T5-base       & & \multicolumn{2}{c}{74.21}       \\\cline{2-5}  
&SD        &  &\multicolumn{2}{c}{79.17}                                                \\
&SD w/HC       &   &\multicolumn{2}{c}{78.89}                                                  \\\cline{2-5}
&Selective SD    &    & \multicolumn{2}{c}{81.02}                                                   \\  
&{PA-KD}    & &\multicolumn{2}{c}{82.56}                                                  \\
\bottomrule
\end{tabular}
\caption{Experimental results of self-distillation (SD) on PTB.}
\label{tab:selfkd}
\end{table}
In addition to employing a capable teacher model, we explore a common scenario where only a single model is accessible. In this scenario, self-distillation (SD)~\cite{Furlanello2018BornAN,mobahi2020sd,liu-etal-2021-noisy} proves to be valuable for enhancing generalization by training the model with self-generated labels. Notably, our proposed Peer-Advised KD is applicable to self-distillation as well. The results in Table~\ref{tab:selfkd} demonstrate that the single-peer PA-KD significantly enhances the performance of the supervised model by approximately 8\% F1, outperforming both direct SD and SD with high-confidence labels.

\end{document}